\documentclass[11pt,a4paper]{article}
\usepackage[hyperref]{emnlp-ijcnlp-2019}
\usepackage{times}
\usepackage{latexsym}
\usepackage{listings}
\usepackage{url}
\usepackage{hyperref,tablefootnote,footnotehyper}
\hypersetup{colorlinks}
\usepackage{algorithm2e}
\usepackage{color, colortbl}
\usepackage{graphicx}

\usepackage{tabularx}
\aclfinalcopy % Uncomment this line for the final submission

\newcommand{\myen}{My$\rightarrow$En~}
\newcommand{\enmy}{En$\rightarrow$My~}

\newcommand{\vc}[1]{\textcolor{blue}{VC: #1}}

\title{Facebook AI's WAT19 Myanmar-English Translation Task Submission}

\author{Peng-Jen Chen*$^1$ \enskip
Jiajun Shen*$^1$ \enskip
Matt Le$^1$ \enskip
Vishrav Chaudhary$^2$ \\
\bf{Ahmed El-Kishky}$^2$ \enskip 
\bf{Guillaume Wenzek}$^1$ \enskip
\bf{Myle Ott}$^1$ \enskip
\bf{Marc'Aurelio Ranzato}$^1$ \\
$^1$Facebook AI Research \enskip $^2$Facebook AI Applied Research \\
{\tt \{pipibjc,jiajunshen,mattle,vishrav,} \\ {\tt ahelk,guw,myleott,ranzato\}@fb.com} \\
}

\date{}

\begin{document}
\maketitle
\begin{abstract}
  This paper describes Facebook AI's submission to the WAT 2019 Myanmar-English translation task \citep{nakazawa-etal-2019-overview}. 
Our baseline systems are BPE-based transformer models.
We explore methods to leverage monolingual data to improve generalization, including self-training, back-translation and their combination.
We further improve results by using noisy channel re-ranking and ensembling. 
We demonstrate that these techniques can significantly improve not only a system trained with additional monolingual data, but even the baseline system
trained exclusively on the provided small parallel dataset. Our system ranks first in both directions according to human evaluation and BLEU, 
with a gain of over 8 BLEU points above the second best system.
\end{abstract}

\renewcommand{\thefootnote}{*}
\footnotetext[1]{Equal contribution.}
\renewcommand\thefootnote{\arabic{footnote}}

\section{Introduction} \label{sec:intro}
% Why we participated: En-My is difficult low resource language pair, multilingual is not applicable, relatively little mono data, languages very distant
% The overall idea of what we did: BT of course, but recent studies show also limitations of BT in low resource settings. Explain why this is applicable here. HEnceforth, we also try ST - and as proposed in this other work we combine the two.
% We find that the besst results are achieved when using mono data and combining ST and BT.
% However, strong performance is achieved even when using just the parallel data.
% Additionally, we use noisy channel and ensembling to further boost results.
% We also report things that did not improve much and were left out from the submission.
While machine translation (MT) has proven very successful for high resource language pairs~\citep{ng2019facebook, msrMT18}, 
it is still an open research question how to make it work well for the vast majority of language pairs which are low resource. In this setting, 
relatively little parallel data is available to train the system and the translation task is even more difficult because the language pairs are 
usually more distant and the domains of the source and target language match less well~\citep{stdm19}.

English-Myanmar is an interesting case study in this respect, because
i) the language of Myanmar is morphologically rich and very different from English, ii) Myanmar language does not bear strong similarities with 
other high-resource languages and therefore does not benefit from multi-lingual training, iii) there is relatively little parallel data available and iv) even 
monolingual data in Myanmar language is difficult to gather due to the multiple encodings of the language.

Motivated by this challenge, we participated in the 2019 edition of the competition on Myanmar-English, organized by the Workshop on Asian Translation. This
paper describes our submission, which achieved the highest human evaluation and BLEU score~\citep{bleu} in the competition.

Following common practice in the field, we used back-translation~\citep{sennrich2015improving} to leverage target side monolingual data.
However, the domain of Myanmar monolingual data is very different from the test domain, which is English 
originating news~\citep{stdm19}. 
Since this may hamper the performance of back-translation, we also explored methods that leverage monolingual data on the source side, which is in-domain with the test set when translating from English to Myanmar.
We investigated the use of self-training~\citep{yarowski, ueffing, zhang16, junxian19} which augments the original parallel data with synthetic data where sources are taken from the
original source monolingual dataset and targets are produced by the current machine translation system. We show that self-training
and back-translation are often complementary to each other and yield additional improvements when applied in an iterative fashion. 

In fact, back-translation and self-training can also be applied when learning from the parallel dataset alone, greatly improving performance over
the baseline using the original bitext data. We also report further improvements by swapping beam search decoding with noisy channel re-ranking~\citep{noisychannel} and by ensembling.

We will start by discussing the data preparation process in \textsection\ref{sec:data}, followed by our model details in \textsection\ref{sec:sys} and results in \textsection\ref{sec:results}. We conclude with some final remarks in \textsection\ref{sec:concl}. In Appendix~\ref{sec:grid_search} we report training details and describe the methods that have not proved useful for this task in Appendix~\ref{sec:unsucc}.

\section{Data} \label{sec:data}

In this section, we describe the data we used for training and the pre-processing we applied. 

\subsection{Parallel Data}
The parallel data was provided by the organizers of the competition and consists of two datasets. The first dataset is the Asian Language Treebank (ALT) corpus~\citep{alt,ding2018nova,ding2019towards} which consists of 18,088 training sentences, 1,000 validation sentences and 1,018 test sentences from English originating news articles. In this dataset, there is a space character separating each Myanmar morpheme~\citep{alt}.

The second dataset is the UCSY dataset\footnote{\url{http://lotus.kuee.kyoto-u.ac.jp/WAT/my-en-data/}} which contains 204,539 sentences from various domains, including news articles and textbooks. The originating language of these sentences is not specified. Unlike the ALT dataset, Myanmar text in the UCSY dataset is not segmented and contains very little spacing as it is typical in this language.

The organizers of the competition evaluate submitted systems on the ALT test set.

We denote the parallel dataset by  $\mathcal{P} = \{X,Y\}$.

\subsection{Monolingual Data}
We gather English monolingual data by taking a subset of the 2018 Newscrawl dataset provided by WMT~\cite{barrault-etal-2019-findings}, which contains approximately 79 million unique sentences.  We choose Newscrawl data to match the domain of the ALT dataset, which primarily contains news originating from English sources.

For Myanmar language, we take five snapshots of the Commoncrawl dataset and combine them with the raw data from~\citet{ngram-commoncrawl}.  After de-duplication, this resulted in approximately 28 million unique lines. This data is not restricted to the news domain.

We denote by $\mathcal{M_S}$ the source monolingual dataset and by $\mathcal{M_T}$ target monolingual dataset.

\subsection{Data Preprocessing}
The Myanmar monolingual data we collect from Commoncrawl contains text in both Unicode and Zawgyi encodings.  We use the \texttt{myanmar-tools}\footnote{\url{https://github.com/google/myanmar-tools}} library to classify and convert all Zawgyi text to Unicode.  Since text classification is performed at the document level, the corpus is left with many embedded English sentences, which we filter by running the fastText classifier~\cite{fasttext} over individual sentences.

We tokenize English text using Moses~\cite{moses} with aggressive hyphen splitting.  We explored multiple approaches for tokenizing Myanmar text, including the provided tokenizer and several open source tools. However, initial experiments showed that leaving the text untokenized yielded the best results. When generating Myanmar translations at inference time, we remove separators introduced by BPE, remove all spaces from the generated text, and then apply the provided tokenizer\footnote{\texttt{myseg.py} can be found in the parallel dataset file on the page \url{http://lotus.kuee.kyoto-u.ac.jp/WAT/my-en-data/}}.

Finally, we use SentencePiece~\cite{sentencepiece} to learn a BPE vocabulary of size 10,000 over the combined English and Myanmar parallel text corpus.  

\section{System Overview} \label{sec:sys}
Our architecture is a transformer-based neural machine translation system trained with \texttt{fairseq}\footnote{\url{https://github.com/pytorch/fairseq}}~\citep{ott2019fairseq}.
We tuned model hyper-parameters via random search over a range of possible values (see Appendix~\ref{sec:grid_search} for details).
We performed early stopping based on perplexity on the ALT validation set, and final model hyper-parameter selection based on the BLEU score on the same validation set.
We never used the ALT test set during development, and only used it for the final reporting at submission time.

Next, we describe several enhancements to this baseline model (\textsection\ref{sec:basic_training}) and to the decoding process (\textsection\ref{sec:nc}).
We also describe several methods for leveraging monolingual data, including our final iterative approach (\textsection\ref{sec:mono}).

\subsection{Improvements to the Baseline Model} \label{sec:basic_training}
% Tagging
% Finetuning
% Ensembling
We improve our baseline neural machine translation system with: tagging~\citep{sennrich16,kobus18,caswell2019tagged}, fine-tuning and ensembling.

\paragraph{Tagging:} Since our test set comes from the ALT corpus and our training set is composed by several datasets from different domains, we prepend to the input source sentence a token specifying the domain of the input data. We have a total of four domain tokens, indicating whether the input source sentence comes from the ALT dataset, the UCSY dataset, the source monolingual data or if it is a back-translation of the target monolingual data (see \textsection\ref{sec:mono} for more details).

\paragraph{Fine-tuning:} The models submitted for final evaluation have also been fine-tuned to the training set of the ALT dataset, as a way to better adapt to the domain of the test set. Fine-tuning is early-stopped based on BLEU on the validation set.

\paragraph{Ensembling:} Finally, since we tune our model hyper-parameters via randomized grid search, we are able to cheaply build an ensemble model from the top $k$ best performing hyper-parameter choices. Ensembling yielded consistent gains of about 1 BLEU point.

\subsection{Improvements to Decoding} \label{sec:nc}
Neural machine translation systems typically employ beam search decoding at inference time to find the most likely hypothesis for a given source sentence.
In this work, we improve upon beam search through noisy-channel reranking~\cite{noisychannel}.
This approach was a key component of the winning submission in the WMT 2019 news translation shared task for English-German, German-English, English-Russian and Russian-English~\citep{ng2019facebook}.
%, and here we follow the same methodology.

More specifically, given a source sentence $x$ and a candidate translation $y$, we compute the following score:
\begin{equation}
\log P(y|x) + \lambda_1 \log P(x|y) + \lambda_2 \log P(y) \label{eq:nc}
\end{equation}
where $\log P(y|x)$, $\log P(x|y)$ and $\log P(y)$ are the forward model, backward model and language model scores, respectively.
This combined score is used to rerank the $n$-best target hypotheses produced by beam search.
In our experiments we set $n$ to 50 and output the highest-scoring hypothesis from this set as our translation.
The weights $\lambda_1$ and $\lambda_2$ are tuned via random search on the validation set. The ranges of values for $\lambda_1$ and $\lambda_2$ are reported in Appendix \ref{sec:grid_search}.

Throughout this work we use noisy channel reranking every time we decode, whether it is to generate forward or backward translations or to generate translations from the final model for evaluation purposes.

Our language models are also based on the transformer architecture and follow the same setup as \citet{radford2018improving}. The English language model is trained on the CC-News dataset \cite{liu2019roberta} and consists of 12 transformer layers and a total of 124M parameters.
The Myanmar language model is first trained on the Commoncrawl monolingual data and then fine-tuned on the Myanmar portion of the ALT parallel training data; it consists of 6 transformer layers and 70M parameters.
For our constrained submission, which does not make use of additional data, we trained smaller transformer language models for each language (5 transformer layers, 8M parameters) using each side of the provided parallel corpus.
For both directions, we observed gains when applying noisy channel reranking, as shown in Table~\ref{tab:noisy_channel}.

\begin{table}[t]
\centering
\small
\begin{tabular}{l r r}
\hline
Model & My$\rightarrow$En & En$\rightarrow$My \\\hline
$\mathcal{P}$ | beam & 25.1 & 35.9\\
$\mathcal{P}$ | reranking & 26.3 & 36.9\\
\hline
$\mathcal{P} \cup \mathcal{M_T}$, beam | beam & 32.2 & 38.8\\
$\mathcal{P} \cup \mathcal{M_T}$, reranking | beam & 32.5 & 38.9\\\hline
$\mathcal{P} \cup \mathcal{M_T}$, reranking | reranking & 35.2 & 39.4 \\\hline
\end{tabular}
\caption{Effect of noisy channel reranking when evaluating on the validation set.
On the left of the "|" symbol there is the dataset used to train the system and the decoding process used to generate back-translated data (if any). On the right of the "|" symbol there is the decoding process used to generate hypotheses from the forward model. $\mathcal{P}$ refers to the parallel dataset and $\mathcal{M_T}$ refers to the target monolingual dataset.
} \label{tab:noisy_channel}
\end{table}

\subsection{Leveraging Monolingual Data} \label{sec:mono}
In this section we describe basic approaches to leverage monolingual data.
Notice however that these methods also improve system performance in the absence of additional monolingual data (i.e., by reusing the available parallel data), see \textsection\ref{sec:sys_with_para}.

We denote by $\overrightarrow{f}$ and $\overleftarrow{g}$ the forward (from source to target) and the backward (from target to source) machine translation systems. 

\paragraph{Back-translation (BT)}~\citep{sennrich2015improving} is an effective data augmentation method  leveraging target side monolingual data. To perform back-translation, we first train $\overleftarrow{g}$ on $\{Y,X\}$ and use it to translate $\mathcal{M_T}$ to produce synthetic source side data, denoted by $\overleftarrow{g}(\mathcal{M_T})$. We then concatenate the original bitext data $\{X,Y\}$ with the back-translated data $\{\overleftarrow{g}(\mathcal{M_T}), \mathcal{M_T}\}$ and train the forward translation model from scratch.
% at a rate between X-Y\mo{TODO fill in}
We typically upsample the original parallel data, with the exact rate tuned together with the other hyper-parameters on the validation set (see Appendix \ref{sec:grid_search} for the upsample ratio range).

\paragraph{Self-Training (ST)}~\citep{ueffing, zhang16, junxian19} instead augments the original parallel dataset $\mathcal{P}=\{X,Y\}$ with synthetic pairs composed by a sentence from the source monolingual dataset with the corresponding forward model translation as target, $\{ (\mathcal{M_S}, \overrightarrow{f}(\mathcal{M_S})\}$.
The potential advantage of this method is that the source side monolingual data can be more in-domain with the test set, which is the case for the English to Myanmar direction. The shortcoming is that synthetic targets are often incorrect and may deteriorate performance.

\paragraph{Combining BT + ST:} Self-training and back-translation are complementary to each other. The former is better when the source monolingual data is in-domain while the latter is better when the target monolingual data is in-domain, relative to the domain of the test set.
%Moreover, self-training trades off the potential gains of learning from in-domain data with the potential loss due to the noise in the synthetic references.

In Table \ref{tab:self_training}, we show that these two approaches can be combined and yield better performance than either method individually.
Specifically, we combine bitext data together with self-trained and back-translated data, $\{X,Y\} \cup \{\overleftarrow{g}(\mathcal{M_T}), \mathcal{M_T}\} \cup  \{ (\mathcal{M_S}, \overrightarrow{f}(\mathcal{M_S})\}$.
As for BT, we upsample the bitext data, concatenate it with the forward and backward translations  and train a new forward model from scratch.
The upsample ratios for each dataset are tuned via hyper-parameter search on the validation set.

\begin{table}[t]
\centering
\begin{tabular}{l r r}
\hline
Model & My$\rightarrow$En & En$\rightarrow$My \\\hline
BT & 33.1 & 39.5\\
ST & 33.2 & 39.9\\
BT + ST & 34.1 & 40.3\\\hline
\end{tabular}
\caption{\label{tab:self_training}
Combining BT and ST yields better BLEU score than BT and ST.}
\end{table}

\begin{algorithm}[t]
\SetAlgoLined
\footnotesize
\nl \KwData{Given a parallel dataset $\{X,Y\}$, a source monolingual dataset $\mathcal{M_S}$ and a target monolingual dataset $\mathcal{M_T}$\;}
\nl Given an initial forward model $\overrightarrow{f}$ and backward model $\overleftarrow{g}$ trained on  $\{X,Y\}$\;
\nl Let $N$ be the number of hyper-parameter configurations evaluated during random search\;
\nl Let $k$ be the number of models used in the ensemble\;
  \nl \For{$t$ \textbf{in} $[1 \dots T]$}{
    \nl forward-translated data: $\mathcal{F} \longleftarrow \overrightarrow{f}(\mathcal{M_S})$\;
    \nl back-translated data: $\mathcal{B} \longleftarrow \overleftarrow{g}(\mathcal{M_T})$\; 
    \nl $\{\overrightarrow{f}_i\}_{i=1 \dots N} \longleftarrow$ random search using: $\{X,Y\} \cup \{\mathcal{M_S}, \mathcal{F} \} \cup \{\mathcal{B}, \mathcal{M_T} \}$\; 
    \nl $\{\overleftarrow{g}_i\}_{i=1 \dots N} \longleftarrow$ random search using: $\{Y,X\} \cup \{\mathcal{F}, \mathcal{M_S} \} \cup \{\mathcal{M_T}, \mathcal{B} \}$\; 
    \nl \If{t == T}{ 
       \nl Fine-tune $\{\overrightarrow{f}_i\}_{i=1 \dots N}$ and $\{\overleftarrow{g}_i\}_{i=1 \dots N}$  on the in-domain ALT dataset\; 
    }
    \nl $\overrightarrow{f}$ $\longleftarrow$ ensemble of top k best models from $\{\overrightarrow{f}_i\}_{i=1 \dots N}$\; 
    \nl $\overleftarrow{g}$ $\longleftarrow$ ensemble of top k best models from $\{\overleftarrow{g}_i\}_{i=1 \dots N}$\; 
  }
 \KwResult{Forward MT system $\overrightarrow{f}$ and backward MT system $\overleftarrow{g}$}
\caption{Iterative Learning Algorithm\label{algo:iterativeSTBT}}
\end{algorithm}

\subsubsection{Final Iterative Algorithm} \label{sec:iterative}
The final algorithm proceeds in rounds as described in Alg.~\ref{algo:iterativeSTBT}. At each round, we are provided with a forward model $\overrightarrow{f}$ and a backward model $\overleftarrow{g}$. The forward model translates source side monolingual data (line 6). This is used as forward-translated data to improve the forward model, and as back-translated data to improve the backward model. Similarly, the backward model is used to back-translate target monolingual data (line 7). This data is then used to improve the forward model via back-translation, but also the backward model via self-training. All these datasets are concatenated and weighted to train new forward and backward models (see lines 8 and 9).
At the very last iteration, models are fine-tuned on the ALT training set (line 11 and 12), and either way, the best models from the random search are combined into an ensemble to define the new forward and backward models (line 13 and 14) to be used at the next iteration.
 This whole process of generation and training then repeats as many times as desired. In our experiments we iterated at most three times.

% Noisy Channel Model Reranking
%\input{noisy_channel.tex}
%\input{finetuning_ensemble.tex}
\section{Results} \label{sec:results}
In this section we report validation BLEU scores for the intermediate iterations and ablations, and test BLEU scores only for  our final submission. Details of the model architecture, data processing and optimization algorithm are reported in Appendix~\ref{sec:grid_search}.

Our baseline system is trained on the provided parallel datasets with the modeling extensions described in \textsection\ref{sec:basic_training}. According to our hyper-parameter search, the optimal upsampling ratio of the smaller in-domain ALT dataset is three and the best forward and backward model have 5 encoder and 5 decoder transformer layers, where the number of attention heads, embedding dimension and inner-layer dimension are 4, 512, 2048, respectively. Each single model is trained on 4 Volta GPUs for 1.4 hours. We refer to this model as the "Baseline" in our result tables.

\begin{table}[t]
    \centering
    \begin{tabular}{l | l | c | c}
        \hline
        & Description & My $\rightarrow$ En & En $\rightarrow$ My \\\hline
        1 & Baseline (single) & 23.3 & 34.9 \\
        2 & \begin{tabular}{@{}l@{}} Baseline \\ (ensemble) \\ \end{tabular} & 25.1 & 35.9 \\
        3 & 2 + reranking & 26.3 & 36.9\\
        4 & 3 + ST & 26.4 & 38.2 \\
        5 & 3 + BT & 26.5 & 36.9 \\
        6 & 3 + (ST + BT) & 27.0 & 38.1\\\hline 
        
    \end{tabular}
    \caption{BLEU scores of systems trained only on the provided parallel datasets.}
    \label{tab:results_para}
\end{table}
\subsection{System Trained on Parallel Data Only}\label{sec:sys_with_para}
We submitted a machine translation system that only uses the provided ALT and UCSY parallel datasets, without any additional monolingual data, results are reported in Tab.~\ref{tab:results_para}. 
The baseline system achieves $23.3$ BLEU points for \myen and $34.9$ for \enmy.
Ensembling 5 models yields +1.8 BLEU points gain for \myen and +1.0 point for \enmy. To apply noisy channel reranking, we train language models {\em using data from the ALT and UCSY training set}. The language model architectures are the same for both languages, each has 5 transformer layers, 4 attention heads, 256 embedding dimensions and 512 inner-layer dimensions. Noisy channel ranking yields a gain of $+1.2$ BLEU points for \myen and $+1.0$ points for \enmy on top of the ensemble models.

To further improve generalization, we also translated the source and target portion of the parallel dataset using the baseline system in order to collect forward-translations of source sentences and back-translations of target sentences. Based on our grid search, we then train a different model architecture than the baseline system, consisting of 4 layers in encoder and decoder, 8 attention heads, 512 embedding dimensions and 2048 inner-layer dimensions. Each model is trained on 4 Volta GPUs for 2.8 hours.
In this case, we train only for one iteration and we ensemble 5 models for each direction followed by reranking.

By applying back-translation and self-training to the parallel data we obtain an additional gain of  $+0.7$ points for \myen and $+1.2$ points for \enmy over the baseline model. We also find that combining back-translation and self-training is beneficial for \myen direction, where we attain an increase of $+0.5$ BLEU compared to applying each method individually. The final BLEU scores on test set are $26.8$ for \myen and $36.8$ for \enmy.

\begin{table}[t]
    \centering
    \begin{tabular}{l | c | c}
        \hline
        Description & My $\rightarrow$ En & En $\rightarrow$ My \\\hline
        Baseline (ensemble) & 25.1 & 35.9 \\
        + reranking & 27.7 & 36.9 \\
        + iter. 1 of ST + BT & 35.5 & 40.1 \\
        + iter. 2 of ST + BT & 36.9 & 40.4 \\
        + iter. 3 of ST + BT & 37.9 & 40.6 \\ \hline
    \end{tabular}
    \caption{BLEU scores of systems trained using additional monolingual data.}
    \label{tab:results_mono}
\end{table}
\subsection{System Using Also Monolingual Data}\label{sec:sys_with_mono}
The results using additional monolingual data are reported in Tab.~\ref{tab:results_mono}.
Starting from the ensemble baseline of the previous section, 
noisy channel reranking now yields a bigger gain for \myen, +2.64 points, since the language model is now trained on much more in-domain target monolingual data.

Using the ensemble and the additional monolingual data, we apply back-translation and self-training for three iterations. For each iteration, we use the best model from the previous iteration to translate monolingual data with noisy channel re-ranking. As before, we combine the original parallel data with the two synthetic datasets, and train models from random initialization. We search over hyper-parameters controlling the model architecture whenever we add more monolingual data.

At the first iteration we back-translate 18M English sentences from Newscrawl and 23M Myanmar sentences from Commoncrawl. The best model architecture has 6 layers in the encoder and decoder, where the number of attention heads, embedding dimension and inner-layer dimension are 1024, 4096, 8, respectively. Each model is trained on 4 Volta GPUs for 17 hours. Ensembling two models for \myen and three models for \enmy strikes a good trade-off between translation quality and decoding efficiency to generate data for the next iteration. The re-ranked ensemble  improves by +7.78 BLEU points for \myen compared to best supervised model, and +3.18 points for \enmy.

At the second iteration, we use the same amount of monolingual data of iteration 1 and repeat the same exact process. The model architecture is the same as in the first iteration. We ensemble two models for \myen and use a single model for \enmy . We further improve upon the previous iteration by +1.41 points for \myen and +0.27 points for \enmy .

At the third and last iteration, we use more monolingual data for both languages, 28M Myanmar sentences and 79M English sentences. We found beneficial~\citep{ng2019facebook} at this iteration to increase FFN dimension to 8192 and the number of heads to 16. Each model is trained on 8 Volta GPUs for 30 hours. After training models on the parallel and synthetic datasets, we fine-tune each of them on the ALT training set, followed by ensembling. We ensemble 5 models for both directions and apply noisy channel re-ranking as our final submission. Compared to iteration 2 models, the final models yield +0.94 points gain for \myen and +0.26 points for \enmy. The BLEU scores of this system on the test set are 38.59 for \myen and 39.25 for \enmy.

\subsection{Final Evaluation}
\begin{table}
    \centering
    \begin{tabular}{l | c | c}
        \hline
        Description (My $\rightarrow$ En) & BLEU & Adequacy \\
        \hline
        \rowcolor{yellow} \bf{FBAI} & 38.6 & 4.4\\
        \rowcolor{yellow} Team1 & 30.2 & 4.0\\
        \bf{FBAI}  & 26.8 & - \\
        \rowcolor{yellow} Team2  & 24.8 & 2.8\\
        Team3 & 19.6 & 1.3\\
        Team4 & 18.5 & - \\
        Team5 & 14.9 & - \\
        \rowcolor{yellow} Team6 & 10.7 & - \\
        \hline 
    \end{tabular}
    \caption{\myen leaderboard\footnotemark. The values are BLEU score (second column) and Adequacy scores (third column). Rows highlighted in yellow identify systems that make use of additional monolingual data. Our system is tagged as FBAI.}
    \label{tab:leaderboard_myen}
\end{table}
\footnotetext[5]{\url{http://lotus.kuee.kyoto-u.ac.jp/WAT/evaluation/list.php?t=70&o=4}}
\begin{table}
    \centering
    \begin{tabular}{l | c | c }
        \hline
        Description (En $\rightarrow$ My) & BLEU & Adequacy \\\hline
        \rowcolor{yellow} \bf{FBAI} & 39.3 & 3.9\\
        \bf{FBAI} & 36.8 & - \\
        \rowcolor{yellow} Team A & 31.3 & 2.4\\
        \rowcolor{yellow} Team B & 30.8 & 2.7\\
        \rowcolor{yellow} Team C & 30.8 & - \\
        \rowcolor{yellow} Team D & 28.2 & - \\
        Team F  & 25.9 & - \\
        Team G & 22.5 & - \\
        \rowcolor{yellow} Team H & 20.9 & 1.1 \\
        Team I & 19.9 & - \\
        \hline 
    \end{tabular}
    \caption{\enmy leaderboard\footnotemark. The values are BLEU score (second column) and Adequacy scores (third column). Rows highlighted in yellow identify systems that make use of additional monolingual data. Our system is tagged as FBAI.}
    \label{tab:leaderboard_enmy}
\end{table}
\footnotetext[6]{\url{http://lotus.kuee.kyoto-u.ac.jp/WAT/evaluation/list.php?t=71&o=9}}
Tables~\ref{tab:leaderboard_myen} and \ref{tab:leaderboard_enmy} report the leaderboard results provided by the organizers of the competition. 
For each direction, they selected the best system of the four teams that scored the best according to BLEU, and they performed 
a JPO adequacy human evaluation~\cite{nakazawa-etal-2018-overview}. 
These evaluations are conducted by professional translations who assign a score between 1 and 5 to each translation based on its adequacy. 
A score equal to 5 points means that all the important information is correctly reported while a score equal to 1 point means that almost all the 
important information is missing or incorrect.

\begin{figure*}[!t]
\begin{center}
\includegraphics[scale=0.4]{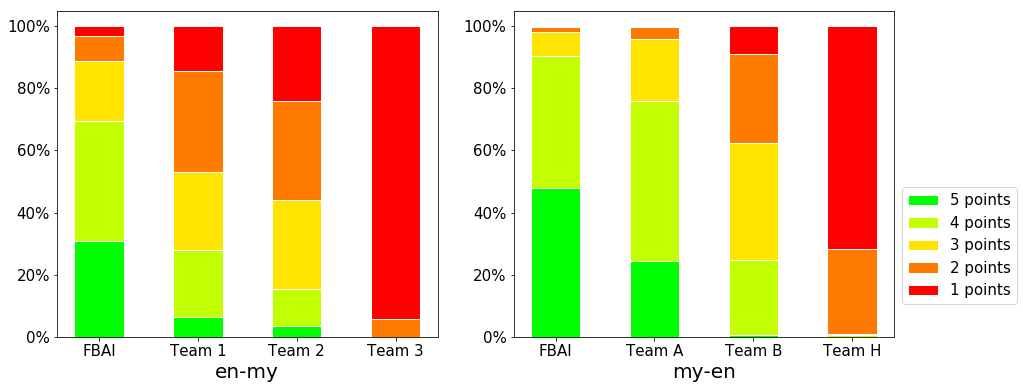}
\end{center}
\caption{\small
Percentage of each adequacy score obtained by the best systems which participated in the competition. Our system is tagged as FBAI.}
\label{fig:adequacy_summary}
\end{figure*}
First, we observe that our system achieves the best BLEU and adequacy score in both directions, with a gain of more than 8 BLEU points over the second best entry for both directions. The average adequacy score is 0.4 point and 1.2 point higher than the second best entry for \myen and \enmy, respectively. Among the rated sentences, more than $30\%$ of sentences translated by our system are rated with 5 points in \enmy, compared to $6.3\%$ of the second best system. For \myen, $48\%$ of our translated sentences are rated with 5 points while the second best system has only $24.5\%$. 
See Fig~\ref{fig:adequacy_summary} for the percentage of each score obtained by the best systems which participated in the competition.

Second, our submission which does not use additional monolingual data is even stronger than all the other submissions in \enmy in terms of BLEU score, including those that do make use of additional monolingual data (see second row of Tab.~\ref{tab:leaderboard_enmy}).

If we consider submissions that only use the provided parallel data (see rows that are not highlighted), our submission improves upon the second best system by 7.2 BLEU in \myen and 10.9 BLEU in \enmy. This suggests that our baseline system is very strong and that applying ST and BT to the parallel dataset is a good way to build even stronger baselines, as demonstrated also in Tab.~\ref{tab:results_para}.

Finally, the gains brought by monolingual datasets is striking only in \myen (+11.8 BLEU points in \myen  compared to only +2.5 BLEU points in \enmy, for our submissions). The reason is because the ALT test set originates from English news and the target English monolingual data is high quality and in-domain with the test set. Moreover, the source originating Myanmar sentences are translationese of English news sentences, a setting which is particularly favorable to BT. Instead, Myanmar monolingual data is out-of-domain and noisy which makes BT much less effective. ST helps improving BT performance as shown in Tab.~\ref{tab:self_training} but the gains are still limited.

\section{Conclusion} \label{sec:concl}
We described the approach we used in our submission to the WAT 2019 Myanmar-English machine translation competition. Our approach achieved the best performance both with and without the use of additional monolingual data.
It is based on several methods which we combine together. First, we use back-translation to help regularizing and adapting to the test domain, particularly in the Myanmar to English direction. Second, we use self-training as a way to better leverage in-domain source-side monolingual data, particularly in the English to Myanmar direction. Third, given the complementary nature of these two approaches we combined them in an iterative fashion. Fourth, we improve decoding by using noisy-channel re-ranking and ensembling. 

We surmise that there is still quite some room for improvement by better leveraging noisy parallel data resources, by better combining together these different sources of additional data, and by designing better approaches to leverage source side monolingual data.

\section*{Acknowledgements}
The Authors wish to thank Sergey Edunov for sharing precious insights about his experience participating in WMT competitions and
Htet Linn for feedback on how spacing is used in Burmese and for checking a handful of translations during early development.

\bibliography{emnlp-ijcnlp-2019}
\bibliographystyle{acl_natbib}
\newpage
\appendix
\section{Hyper-Parameter Search} \label{sec:grid_search}
In this section we report the set of hyper-parameters and range  of values that we used in our random hyper-parameter search. 
For each experiment we searched using $N=30$ hyper-parameter configurations.

Notice that the actual range of hyper-parameters searched in each experiment may be smaller than reported below; for instance, if a model shows signs of overfitting we may search up to 5 layers as opposed to 6 at the next iteration.

\begin{itemize}

\item Layers: \{4, 5, 6\}

\item Embedding dim: \{128, 256, 512, 1024\}

\item FFN dim: \{128, 256, 512, 1024, 2048, 4096, 8192\}

\item Attention heads: \{1, 2, 4, 8, 16\}

\item Dropout: \{0.1, 0.2, 0.3, 0.4, 0.5\}

\item Batch size (number of tokens): \{1, 2, 4, 8, 12, 16, 24, 32\} (multiply by 16000)

\item Label smoothing: \{0.1, 0.2, 0.3\}

\item Learning rate: \{1, 3, 5, 7, 10, 30, 50, 100, 300, 500\} (multiply by 1e-4)

\item Seed: \{1, 2, 3, ... , 30\}

\item Data upsampling ratio
    \begin{itemize}
      \item bitext: \{1, 2, 3, 4, 6, 8, 12, 16, 20, 32, 40, 64\}
      \item forward-translated: \{1, 2, 3, 4, 6, 8, 9\}
      \item back-translated: \{1, 2, 3, 4, 6, 8, 9\}
    \end{itemize}
    
\end{itemize}

When applying noisy-channel reranking, we tune the hyper-parameters $\lambda_1$ and $\lambda_2$ on the validation set. The ranges of the two hyper-parameters are between 0 and 3.

\section{Things We Tried But Did Not Use} \label{sec:unsucc}
This section details attempts that did not significantly improve the overall performance of our translation system and which were therefore left out of the final system. 

\subsection{Out-of-domain parallel data} \label{sec:out_of_domain_para}
Similarly to~\citet{flores2019} we added out-of-domain parallel data from various sources of the OPUS  repository\footnote{\url{http://opus.nlpl.eu/}}, namely GNOME/Ubuntu, QED and GlobalVoices. This provides an additional 38,459 sentence pairs. We also considered two versions of Bible translations from the bible-corpus\footnote{\url{https://github.com/christos-c/bible-corpus/}} resulting in additional 61,843 sentence pairs. Adding this data improved the baseline system by +0.17 BLEU for \myen and +0.26 BLEU for \enmy.

\subsection{Pre-training}
We pre-trained our translation system using a cross-lingual language modeling task~\citep{xlm} as well as a Denoising Auto-Encoding (DAE) task~\citep{dae}. They both did not provide significant improvements; in the following, we report our results using DAE. 

In this setting, we have a single encoder-decoder model which takes a batch of monolingual data, encodes it with the model's encoder, prepends the encoded representation with a language-specific token, and then tries to reconstruct the original input using the model's decoder.  
%This process is iteratively applied to Myanmar then English monolingual data until model convergence \mr{you do not shuffle?}.  
Additionally, the source sentences are corrupted using three different types of noise: word dropping, word blanking, and word swapping~\citep{unsupNMTlample, lample_emnlp2018}.  The goal is to encourage the model to learn some kind of common representation for both languages. 

We found some gains, particularly for the \enmy direction, however, doing backtranslation on top of DAE pretraining did worse or did not improve compared to backtranslation without DAE pretraining.  For this reason, we decided to leave this technique out of our final system. 

\subsection{PBSMT}
We also train a phrase based system using Moses with a default setting. We preprocessed the data using moses tokenizer for English sentences. For  Myanmar sentences, we use BPE instead. We train a count-based 5-gram English and Myanmar language models on the monolingual data we collect. We tune the system using MERT on the ALT validation set. However, the phrase based system does not perform as good as our NMT baseline. The phrase based system we train on the parallel data only yields 10.98 BLEU for \myen and 21.89 BLEU for \enmy, which are 12.32 and 13.05 BLEU points lower than our supervised single NMT model. 

\subsection{Weak Supervision}
For augmenting the original training data with a noisy set of parallel sentences, we mine bitexts from Commoncrawl. This is achieved by first aligning the webpages in English and Myanmar and then extracting parallel sentences from them.
To align webpages, we perform sentence alignment using the IBM1 sentence alignment algorithm~\cite{Brown:1993}, trained on the provided parallel data to obtain bilingual dictionaries from English to Myanmar and Myanmar to English. Using these dictionaries, unigram-based Myanmar translations are added to the English web documents and Myanmar translations are added to the English documents. The similarity score of a document pair $a$ and $b$ is computed as:
\begin{equation}
    sim(a,b) = Lev(url_a, url_b) \times Jaccard(a,b)
\end{equation}
where $Lev(url_a,url_b)$ is the Levenshtein similarity between the $url_a$ and $url_b$ and $Jaccard(a,b)$ is the Jaccard similarity between documents $a$ and $b$. Finally, a one-to-one matching between English and Myanmar documents is enforced by applying a greedy bipartite matching algorithm as described in~\citet{buck2016quick}. The set of matched aligned documents is then mined for parallel bitexts.

We align sentences within two comparable webpages by following the methods outlined in the parallel corpus filtering shared task for low-resource languages~\cite{filtering:2019:WMT}. One of the best performing methods for this task used the LASER model~\cite{artetxe2018multi} to gauge similarity between sentence pairs~\cite{wmt19-filtering-facebook}. Since the open-source LASER model is only trained with 2,000 Myanmar-English bitexts, we retrained the model using the provided UCSY and ALT corpora. For tuning, we use similarity error on the ALT validation dataset and observe that the model performs rather poorly as the available training data was substantially lower than the original setup. 
\iffalse
We also experimented with an alternative method of aligning sentences using the IBM1 scores. The IBM1 scores are computed similar to \citet{Brown:1993} as shown below:

\begin{equation}
p(t|s) = \frac{1}{(l+1)^m}\prod_{j=1}^{m}\sum_{i=0}^{l}p(t_j|s_i)
\end{equation}

where $t$ is the target, $s$ is the source and $l$ and $m$ denote the length of the source and the target sentences, respectively. For improving the quality of sentence pairs scored using IBM1 scores, we added a penalty based on common unigrams and punctuations between the source and the target sentence. We also mined sentences from Myanmar and English Wikipedia aligned documents using this technique. Adding mined bitexts from Commoncrawl and Wikipedia resulted in +X BLEU over the supervised baseline. \vc{I had computed the result separately for CC and Wiki and it had resulted in ~+1.5 BLEU for en-my for both the corpora. I am now rerunning the experiments with combination of both these corpora and will update the results.}
\fi

\end{document}